\documentclass[letterpaper]{article} 
\usepackage[submission]{aaai24}  
\usepackage{times}  
\usepackage{helvet}  
\usepackage{courier}  
\usepackage[hyphens]{url}  
\usepackage{graphicx} 
\usepackage{natbib}  
\usepackage{caption} 
\usepackage{algorithm}
\usepackage{algorithmic}
\usepackage{svg}
\usepackage{subcaption}

\usepackage{amsfonts} 

\usepackage{newfloat}
\usepackage{listings}

\usepackage{amsmath}
\DeclareCaptionStyle{ruled}{labelfont=normalfont,labelsep=colon,strut=off} 
\floatstyle{ruled}
\newfloat{listing}{tb}{lst}{}
\floatname{listing}{Listing}
\title{Online Learning of Human Constraints from Feedback in Shared Autonomy}
\author {
    Shibei Zhu\textsuperscript{\rm 1},
    Tran Nguyen Le\textsuperscript{\rm 2},
    Samuel Kaski\textsuperscript{\rm 1,3},
    Ville Kyrki\textsuperscript{\rm 2}
}
\affiliations {
    \textsuperscript{\rm 1}Department of Computer Science, Aalto University, Finland.\\

    \textsuperscript{\rm 2}Department of Electrical Engineering and Automation, Aalto University, Finland.\\
    \textsuperscript{\rm 3}Department of Computer Science, University of Manchester, United Kingdom\\
}

\usepackage{bibentry}

\begin{document}
\maketitle

\begin{abstract}

Real-time collaboration with humans poses challenges due to the different behavior patterns of humans resulting from diverse physical constraints. Existing works typically focus on learning safety constraints for collaboration, or how to divide and distribute the subtasks between the participating agents to carry out the main task. In contrast, we propose to learn human constraints model that, in addition, consider the diverse behaviors of different human operators. We consider a type of collaboration in a shared-autonomy fashion, where both a human operator and an assistive robot act simultaneously in the same task space that affects each other's actions. The task of the assistive agent is to augment the skill of humans to perform a shared task by supporting humans as much as possible, both in terms of reducing the workload and minimizing the discomfort for the human operator. Therefore, we propose an augmentative assistant agent capable of learning and adapting to human physical constraints, aligning its actions with the ergonomic preferences and limitations of the human operator. 

\end{abstract}
\section{Introduction}

Collaboration forms an essential part of our daily life. 
In recent years, technological advances in both artificial intelligence and robotics allowed us to use robots to empower humans to perform repetitive tasks or physically demanding tasks~\cite{nemec2013transfer,varier2020collaborative,clegg2020learning}. However, designing an assistive robot that takes into account human demand and adapts its policy to different humans in real-time is challenging. Different individuals present with different physical capabilities and even changing capabilities during the collaboration, i.e., due to fatigue levels. The assistant should not only take into account how to collaborate with humans safely, but also consider their physical constraints or capability on the fly during the collaboration. While some physical aspects of humans can be defined, such as their height or muscle strength, some other factors are subject to changes, such as the fatigue level of the individuals. These factors vary from individual to individual. Thus, it is difficult to establish some deterministic rules or mathematical forms for the assistive agent that account for all the situations during the interaction. Designing an adaptive agent that detects different human constraints in real-time is essential in order to deliver a more satisfactory collaboration experience and quality of collaboration. 

\begin{figure}[t]
    \centering
    \includegraphics[width=\columnwidth]{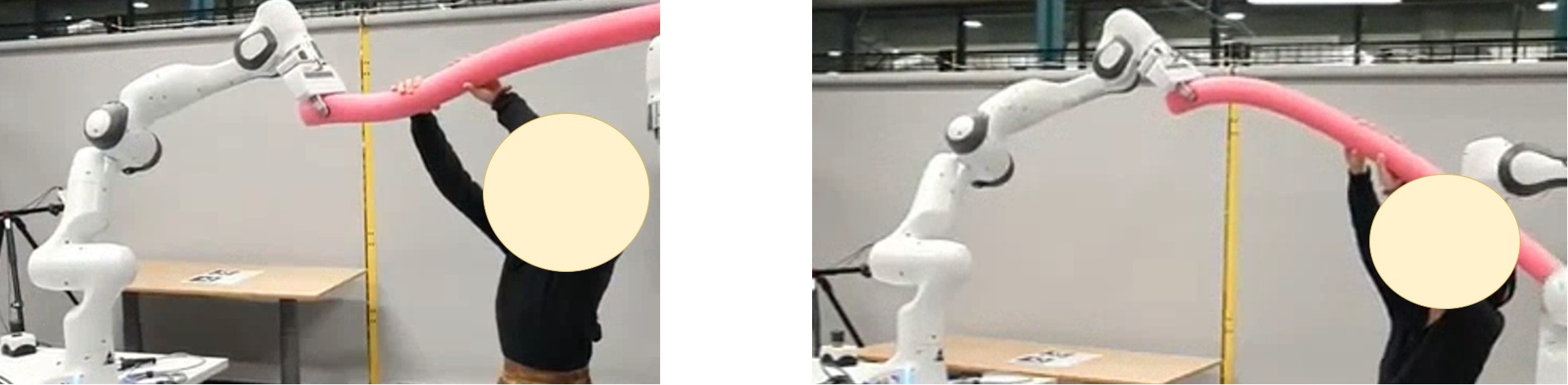}
    \caption{Co-transportation task where both human and robot operate on the same object. \textbf{Left}: assistive agent that considers human physical constraints. \textbf{Right}: assistive agent without considering human's physical constraints, and human operator tries to retake control by pulling back the object.}
    \label{fig:example}
\end{figure}
To date, Reinforcement learning (RL), specifically multi-agent reinforcement learning (MARL), has shown enormous success in collaborative tasks~\cite{kok2006collaborative, nikolaidis2015efficient,omidshafiei2017deep,wang2018facilitating}. Given a shared goal, multi-agent reinforcement learning allows us to formulate a collaborative task as a Markov Game where an assistive agent can learn how to collaborate with another agent by jointly maximizing their shared task reward. Most of the existing works in collaborative tasks only focus on how to distribute a single task into subtasks such that the task load can be divided among different participants in the task \cite{kwon2011towards,wu2021too,chen2020trust}. However, when it comes to a human-robot collaborative task setup, the collaboration typically consists of how to transfer autonomy smoothly between a robot agent and a human operator back and forth in order to complete a given task. Essentially, this means that each agent has its own subtasks or tasks space where the collaboration takes place only when they need to transfer the autonomy. In this paper, we address another type of collaboration task, where both the human and the assistive agent operate in the same task space with the same shared goal with shared autonomy. In such a scenario, the agent needs to learn the human constraints online in order to adapt its policy so that their joint actions are physically feasible. To the knowledge of the authors, the problem of online constraint learning from human feedback remains underexplored, nor its application in a collaborative physical human-robot interaction framework.

To address the aforementioned open issues, we propose an adaptive agent that learns different human constraints as the human model from the human feedback during the physical interaction. This allows the agent to adapt its policy in real-time that accommodate different human needs. 


\section{Problem formulation}
We consider a type of collaborative task in which the robot assists a human operator to perform the task by augmenting the human operator's skills. Let us consider a multi-agent human-robot collaborative framework formulated as a two-player Markov Game $\mathcal{M} = \langle  \mathcal{S}, \mathcal{A}^H, \mathcal{A}^R, p, p_0,  R,\gamma, T \rangle $, where $\mathcal{S} \in \mathbb{R}^m$ is the state space, $\mathcal{A}^H \in \mathbb{R}^n$ and $\mathcal{A}^R \in \mathbb{R}^n $ are the robot and human's respective action spaces, subject to different physical constraints. The joint action space is defined as $\mathcal{A} = \mathcal{A}^H \times \mathcal{A}^R$, and the state transition dynamic is given as $p: \mathcal{S}\times \mathcal{A} \times \mathcal{S} \rightarrow [0, 1]$ with $p_0$ as the initial state distribution. And $R$ is the shared task reward. The joint Q function that defines the expected return for both agents is defined as $Q(s, a^R, a^H) =  \mathbb{E} \Big[ \sum_{t=0}^{T-1} \gamma^{t} R_{\textbf{t}}( s_t, a_t^R, a_t^H;\pi_H, \pi_R ) \Big]$

Both robot and human actions are constrained by their own physical limitations, denoted as $\mathcal{C}_R$ and $\mathcal{C}_{H_{\theta}}$ respectively. The former is explicitly given by its corresponding joint limitations. However, the latter remains unknown and subject to each individual. 
Human policies might not necessarily be the result of rational decisions, but rather noisy, often referred to as Boltzmann rational~\cite{ziebart2008maximum, boularias2011relative, malik2018efficient}. Thus, their corresponding policy that optimizes their own constraint while taking into account robot's physical constraints is given as:

\begin{align}
        \pi_{H}^*(a^H \vert a^R, s) =  & \frac{e^{Q(s, a^H, a^R)}/\beta }{\int_{a^H} e^{Q(s, a^R, a^H)}/\beta}  \\
      \operatorname{subject \;to} \; &\pi_{H} (a^H|s) \leq \mathbf{C}_{{H}_\theta}, \\
          a^{R} & \leq \mathbf{C}_{R} 
    \label{eq:h_constraint}
\end{align}
where $\beta$ is the temperature parameter that adjusts the randomness or the rationality of the decision. 

\begin{figure}
    \centering
    \includegraphics[width =.7\columnwidth]{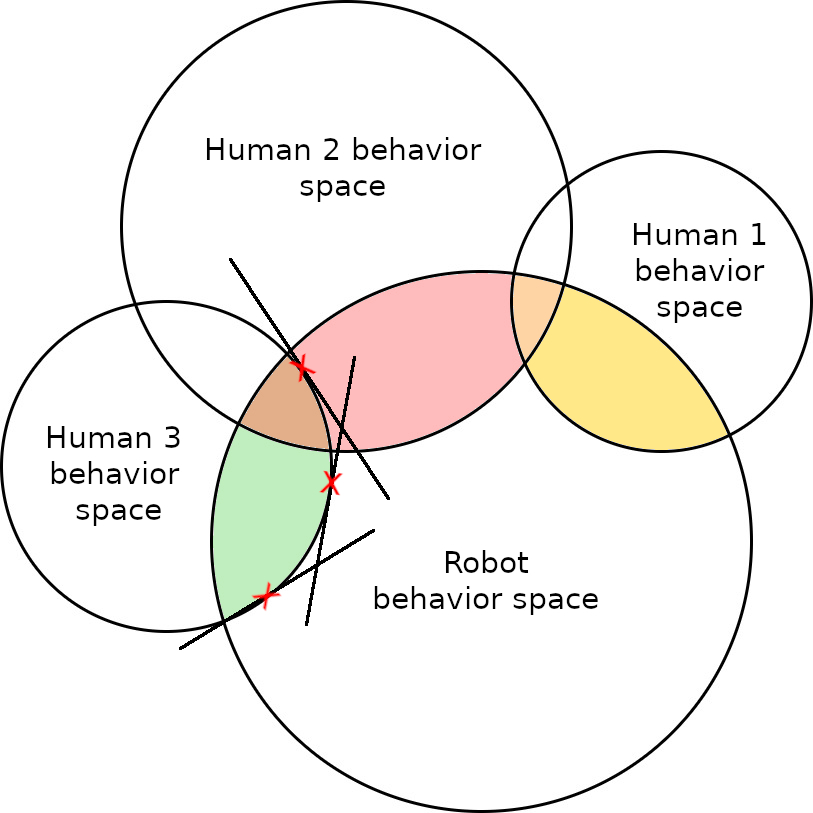}
    \caption{Constraints in the task space where different agents have their constraints in the action space. Given the behavior spaces of humans and robots constrained by their respective physical constraints, the colored areas represent the trust region that defines the plausible set of joint behavior that satisfies all the constraints. We represent the lower bound of the trust region as the constraints that define the upper bound of the human behavior space. }
    \label{fig:behave_space}
\end{figure}

As human constraints cannot be universally defined and consider all individuals, we propose to construct a feasible joint-action region based on the external feedback during the collaboration. This is analogue to a \textbf{trust region} $\mathcal{A}_{\theta}$ that defines the valid joint-action space shared by the human and the assistive agent, where both agents' physical constraints are satisfied. The upper bound of this trust region is defined by the physical constraints of the robot and its lower bound is given by different human constraints as shown in Fig~\ref{fig:behave_space}. Mathematically, this region is defined as:
\begin{equation}
    A_\theta = \{ (a^R, a^H) \in \mathbb{R}^n :  \mathbf{C}_{{H}_{\theta}} \leq (a^R, a^H) \leq \mathbf{C}_R  \}
    \label{eq:trust_region}
\end{equation}

Human physical constraints can be influenced by various sources of factors, i.e., physical limitations, fatigue levels, etc. Therefore, it is not trivial to establish deterministic values to define these constraints for different humans under different circumstances, but rather be defined as a feasible region of joint actions.
We can see an illustrative example in Fig~\ref{fig:behave_space}: the assistive robot and different human operators have different physical constraints that define different behavior spaces. Ideally, the assistive agent should adapt its policy considering human physical constraints such that interaction will not violate their respective physical constraints. 
Thus, the desired assistive agent's policy is re-written as:


\begin{align}
        \pi_{R}^*(a^{R} | s, a^{H},\mathcal{A}_{\theta} ) = \frac{e^{Q(s, a^H, a^R)}/\beta }{\int_{a^R} e^{Q(s, a^R, a^H)}/\beta} \\
      \operatorname{subject \;to} \;   (a^{H}, a^R) & \in \mathcal{A}_{\theta} \label{eq:r_constraint_2}
\end{align}


\section{Learning a trust region from real-time feedback with human constraint model}

As discussed previously, it is hard to establish some deterministic values for human constraints, as many factors would influence these values. Instead of learning some fixed values, we propose a human constraint model to learn a trust region where the joint actions are compatible for both the robot and the current human operator. The upper bound of this trust region is given by the physical constraints of the assistive agent, whereas human constraints draw the lower bound of this region. Our goal is to define the feasible joint-action region by finding this lower bound based on human external feedback. 

\paragraph{Human's feedback as indirect labels for the negative and positive samples.} 
Humans can influence the action of the assistive agent by applying a wrench $w_h \in \mathbb{R}^6$ to deviate the agent from its current predicted action. 
A wrench is a six-dimensional vector consisting of a force $f_h \in \mathbb{R}^3$ and a torque $\mathcal{\tau}_h \in \mathbb{R}^3$. We assume that this additional wrench is applied when a human reaches the upper bound of their physical constraints. In the simplest case, the assistive agent can use its sensory information and the current observation of human policy to determine whether its current policy falls out of the trust region, as follows:

\begin{equation}
   (\pi_R \in \text{trust region} |w_h)  =\begin{cases}
 \text{positive}, & \text{if}\; w_h > \delta\\
\text{negative} , &\text{otherwise}   \\
\end{cases}
\label{eq:feedback}
\end{equation}
Where $\delta$ is a threshold that represents the minimum external feedback value when a human provides any feedback or collaborates in the task. Any value reading below this threshold indicates that no human action is detected and a possible scenario when they are out of their physically operable region that prevents them from collaborating.

Assume that human operators would take actions that bring the joint action back to their comfort zone or feasible region; this feedback information allows us to define the trust region $A_\theta$ in Eq.~\ref{eq:trust_region}. We propose to estimate the unknown lower bound of this trust region $ \mathbf{C}_{{H}_{\theta}}$ based on the human's external forces. In the simplest case, we have linear constraints that define the surface or a hyperplane that separates the positive and negative samples of the trust region, as shown in Fig.~\ref{fig:behave_space}. We define a constraint model to predict human constraints to define hyperplane based on feedback as described in Eq.~\ref{eq:feedback}:
\begin{equation}
    \mathbf{C}_{{H}_{\theta}} = f(a^R, a^H, s, w_h ; \theta)
\end{equation}

\begin{figure}
    \centering
    \includegraphics[width =0.9\columnwidth]{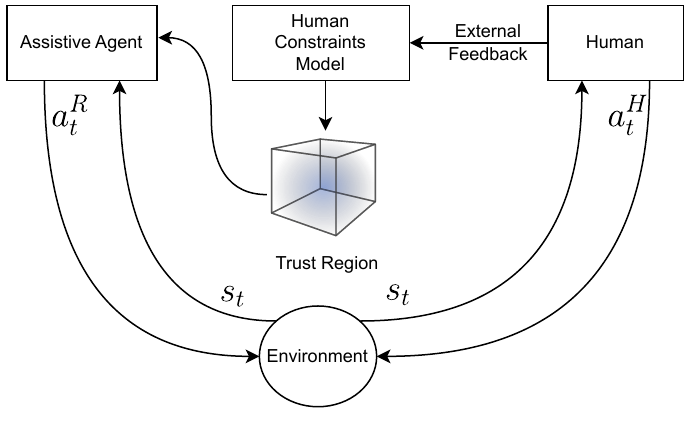}
    \caption{Given the external feedback received in real-time from a human, we develop a human constraint model to define a lower bound of the trust region where the joint actions of both the human and the assistive agent satisfy their constraints. }
    \label{fig:pol_agent}
\end{figure}

The overall architecture can be found in Fig.~\ref{fig:pol_agent}. Where the loss function can be defined based on the feedback as the binary cross-entropy loss using Eq.~\ref{eq:feedback}.




\section{Potential use cases}
\begin{figure}
    \begin{subfigure}{\linewidth}
		\centering
        \includegraphics[width=.45\linewidth]{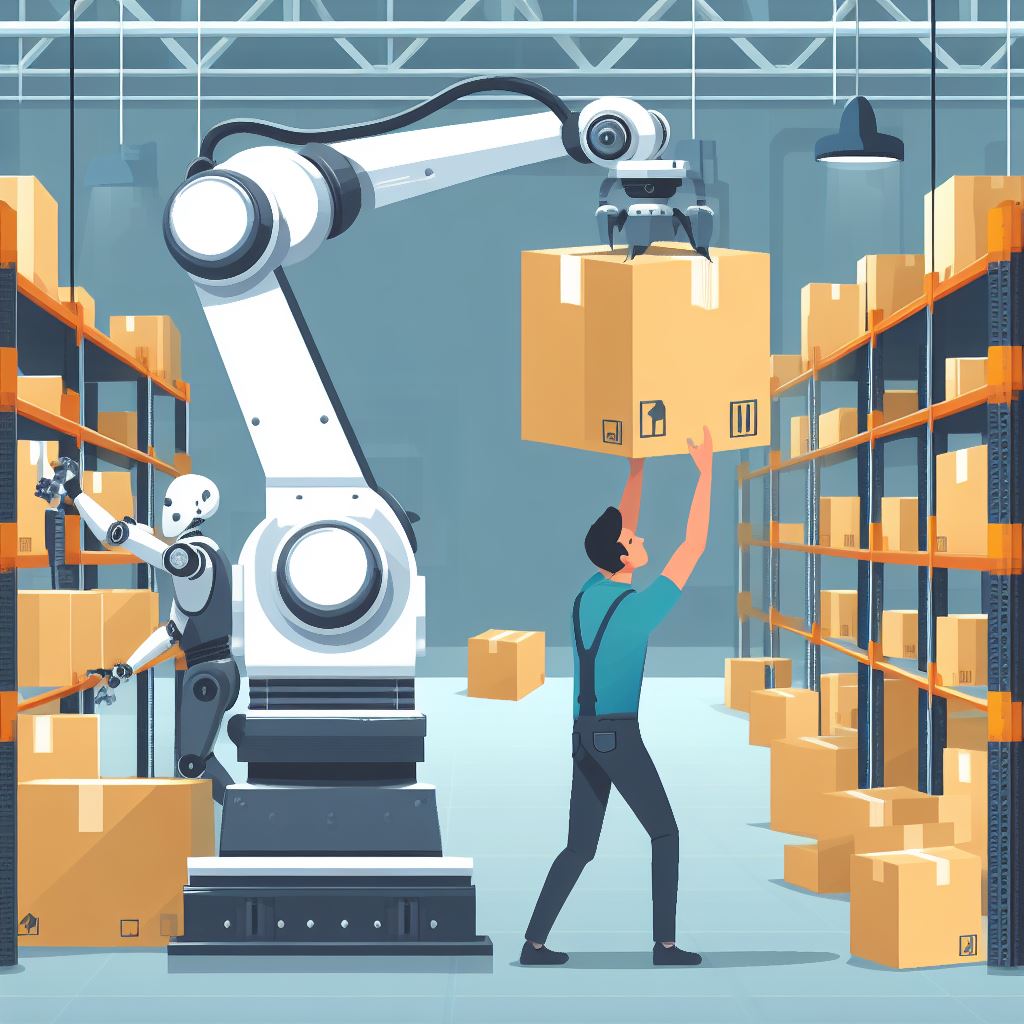}
        \hspace{1em}
        \includegraphics[width=.45\linewidth]{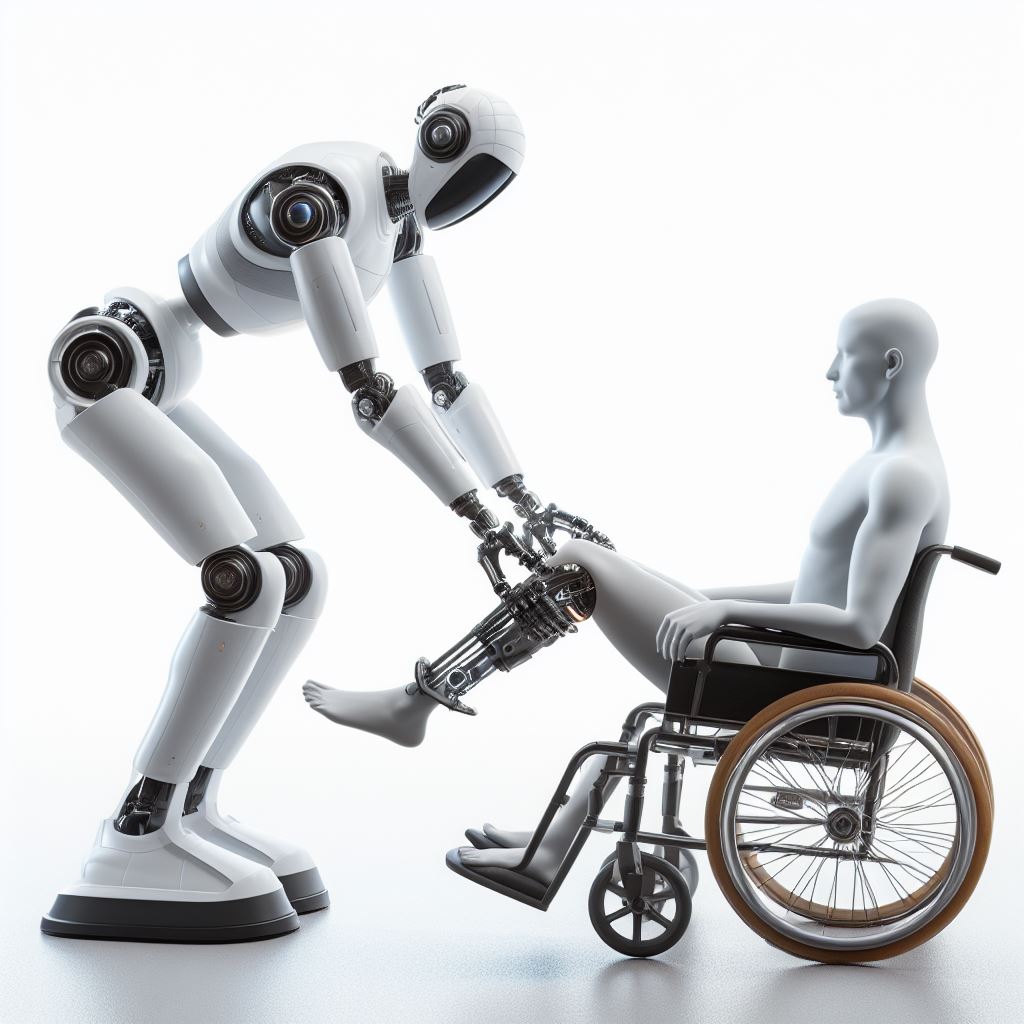}
	\end{subfigure}
    \caption{Robot assists human in co-transportation task (\textbf{Left}) and rehabilitation (\textbf{Right}).}
    \label{fig:cotransportation}
\end{figure}

\paragraph{Co-transportation.} One of the most obvious use cases for the proposed method is the co-transportation task where a robot assists a human in transporting a heavy object to a designated location, as shown in Fig~\ref{fig:cotransportation}. Within this scenario, humans usually experience fatigue due to various factors, such as the heavy weight of the object or uncomfortable lifting poses. Ideally, the proposed method should empower the robot to understand these human constraints, thereby bending its policy to alleviate the physical strain on the human while efficiently achieving the task at hand.

\paragraph{Rehabilitation robots.} In addition to co-transportation, other user cases can be found in medical domains. Examples include rehabilitation, as shown in Fig~\ref{fig:cotransportation}, where a rehabilitation robot or therapeutic robots assist people with manipulative disabilities~\cite{tejima2001rehabilitation}. In such cases, the robot is equipped with actuation to move patients' limbs in order to compensate for the physical capabilities of the patients. The main task of the assistive robot is to progressively aid patients through the rehabilitation process to recover their normal daily functions~\cite{qian2015recent}. Thus, a human constraint model is needed to detect and adapt different strategies during the rehabilitation process. 

\paragraph{Exoskeleton robots.} Similar cases can be found with exoskeleton robots~\cite{veneman2007design, zhang2017human}, where a personalized assistive robot is needed to augment different human operators with different physical capabilities in real time to perform some challenging tasks. Similar to the previous case, the assistive agent augments the mobility of a human by taking into account the physical limitations of different patients in order to augment their physical capabilities. For instance, in~\cite{ivaldi2021using} exoskeleton is used to assist medical staff with heavy-load tasks such as lifting patients. Their study shows that different individuals have different preferences toward different exoskeletons, where improvement can be introduced by adding an adaptive assistive robot that accommodates different individual needs.

\section{Experiments}
\begin{figure*}[t]
    \centering
    \includegraphics[width=.9\textwidth]{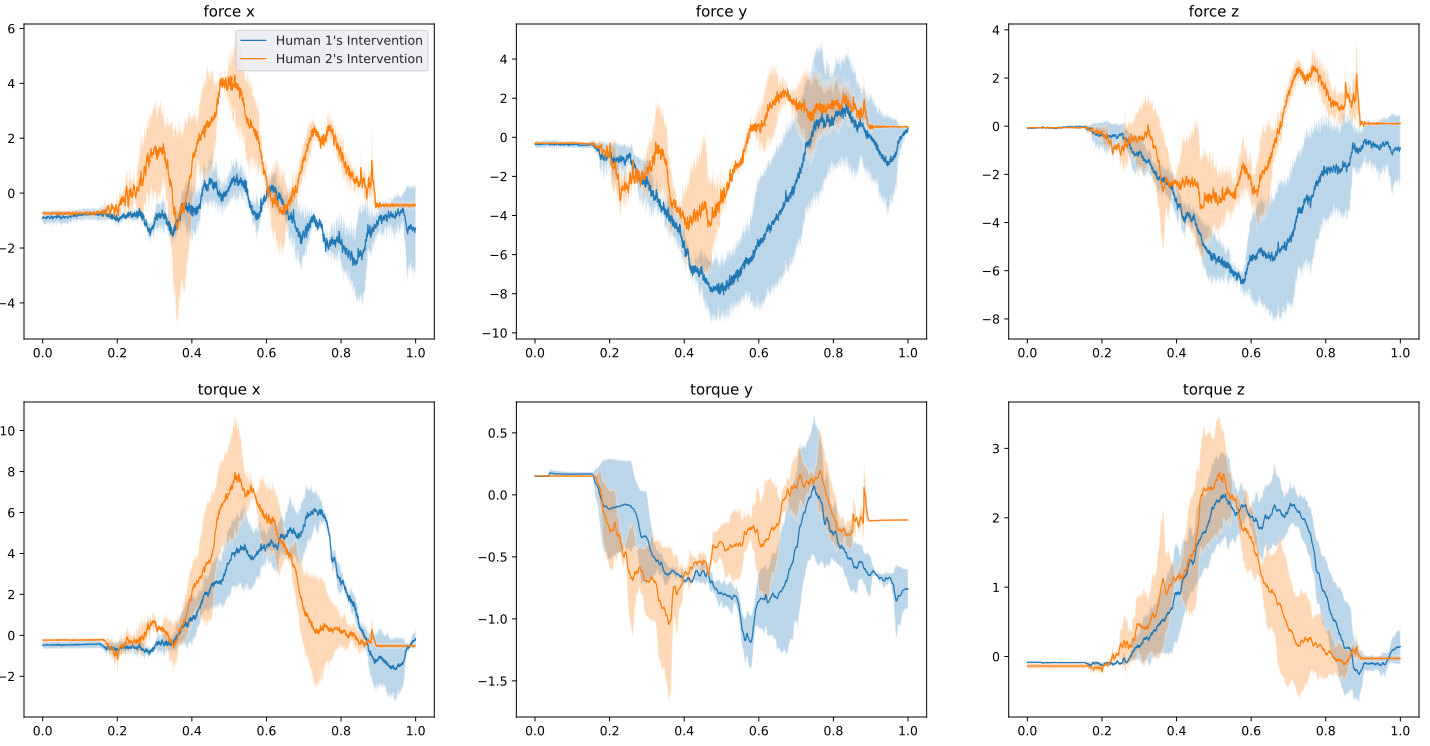}
    \caption{Real human feedback from 2 individuals during the collaborative tasks, represented by their means and standard deviations. The force feedback is six-dimensional vector that is represented by each plot. The x-axis represents the time, and the y-axis the corresponding value reading. 5 runs are used to generate these plots.}
    \label{fig:ft_plot}
\end{figure*}
As mentioned earlier, co-transportation is one of the most obvious use cases for the proposed method. Therefore, to gain a deeper insight into the challenges in this task setting, we conducted an experiment in which a human and a robot collaborated to transport a heavy object from point A to point B. 

In this experiment, the robot's policy was executed in a compliant manner, allowing the human to exert influence on the robot's configuration. This experiment was systematically repeated with two different human operators, each performing the task five times. The external forces and torques that humans acted on the robot are plotted in Fig~\ref{fig:ft_plot}. From the plot we can see that two different humans yield two different sets of external forces and torques, signifying distinct physical constraints imposed by individual strengths and attributes, such as height. Furthermore, the measurements also align with the problem illustrated in Fig~\ref{fig:behave_space}. It is also worth mentioning that the feedback from the same individual yields different variations, which supports our argument that humans are not always optimal decision-makers, but instead subject to some degree of noise.







\bigskip

\section{Conclusion}
In this paper, we propose a new learning framework that allows us to learn human constraints that are diverse and subject to different circumstances in a collaborative physical human-robot setup. While existing works focus on learning safety constraints offline, we propose learning human physical constraints online with human feedback in a collaborative task environment. Unlike the existing works on collaborative tasks that focuses mostly on how to transition autonomy from one agent to another, our learning agent focuses on adaptively learning human physical constraints during the interaction to augment human skills.
\bibliography{references}

\end{document}